\documentclass[10pt,twocolumn,letterpaper]{article}

\usepackage[pagenumbers]{cvpr}      
\usepackage{tikz}
\usetikzlibrary{calc}
\usepackage{pifont}            
\newcommand{\cmark}{\ding{51}} 
\newcommand{\xmark}{\ding{55}} 
\usepackage{verbatim}
\usepackage{booktabs}
\usepackage{makecell}
\usepackage{diagbox}
\usepackage{tcolorbox}
\usepackage{mdframed}
\usepackage{amsmath}
\usepackage{multirow}
\DeclareMathOperator*{\argmax}{argmax}
\usepackage{fancyhdr}
\fancypagestyle{firstpage}{
  \fancyhf{}
  \lfoot{\small Preprint.\ Under review.}

}

\newtheorem{definition}{Definition}[section]
\definecolor{cvprblue}{rgb}{0.21,0.49,0.74}
\usepackage[pagebackref,breaklinks,colorlinks,allcolors=cvprblue]{hyperref}

\def\paperID{xxx} 
\def\confName{CVPR}
\def\confYear{2026}

\title{MMR-Bench: A Comprehensive Benchmark for Multimodal LLM Routing}

\author{
Haoxuan Ma$^{1,2\dagger}$,
Guannan Lai$^{1,2\dagger}$,
and Han-Jia Ye$^{1,2}$\thanks{$^\dagger$Equal contribution. $^*$Corresponding author.}\\
$^{1}$School of Artificial Intelligence, Nanjing University\\
$^{2}$National Key Laboratory for Novel Software Technology, Nanjing University\\
\textit{Email:} \{mahx, laign, yehj\}@lamda.nju.edu.cn
}

\begin{document}
\maketitle
\thispagestyle{firstpage}

\begin{abstract}
Multimodal large language models (MLLMs) have advanced rapidly,  yet heterogeneity in architecture, alignment strategies, and efficiency means that no single model is uniformly superior across tasks. In practical deployments, workloads span lightweight OCR to complex multimodal reasoning; using one MLLM for all queries either over-provisions compute on easy instances or sacrifices accuracy on hard ones. Query-level model selection (routing) addresses this tension, but extending routing from text-only LLMs to MLLMs is nontrivial due to modality fusion, wide variation in computational cost across models, and the absence of a standardized, budget-aware evaluation.

We present \textbf{MMR-Bench}, a unified benchmark that isolates the multimodal routing problem and enables comparison under fixed candidate sets and cost models. MMR-Bench provides (i) a controlled environment with modality-aware inputs and variable compute budgets, (ii) a broad suite of vision–language tasks covering OCR, general VQA, and multimodal math reasoning, and (iii) strong single-model reference, oracle upper bounds, and representative routing policies. Using MMR-Bench, we show that incorporating multimodal signals improves routing quality. Empirically, these cues improve the cost-accuracy frontier and enable the routed system to exceed the strongest single model’s accuracy at roughly \(33\%\) of its cost. Furthermore, policies trained on a subset of models and tasks generalize zero-shot to new datasets and text-only benchmarks without retuning, establishing MMR-Bench as a foundation for studying adaptive multimodal model selection and efficient MLLM deployment. \textit{The code will be available at: \url{https://github.com/Hunter-Wrynn/MMR-Bench}.}

\end{abstract}
    
\section{Introduction}
\label{sec:intro}
\begin{figure}[htbp]
\centering
  \includegraphics[width=\linewidth]{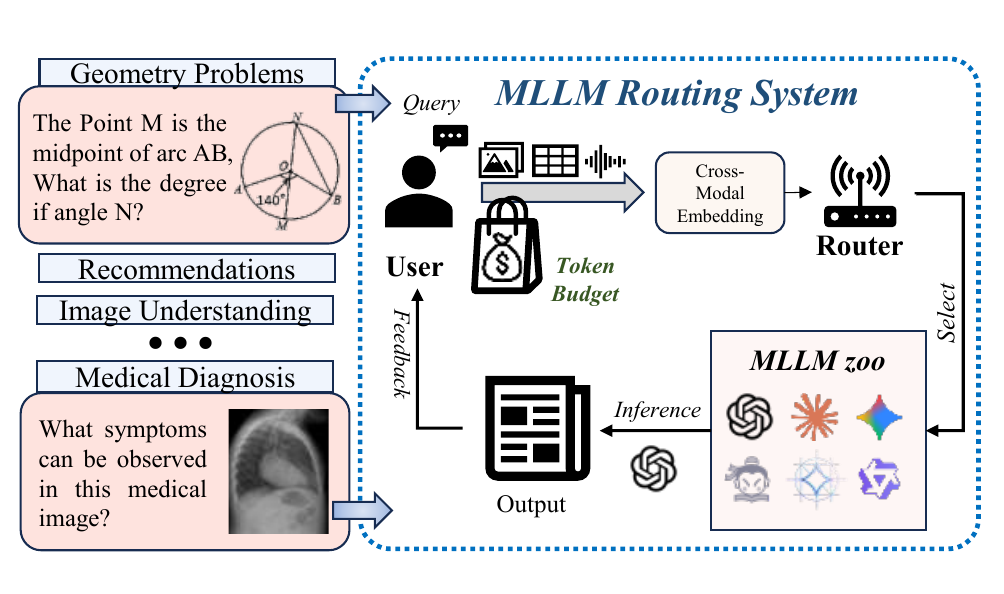}
\caption{The workflow of MLLM routing. It can be applied to various tasks such as mathematical reasoning, multimodal recommendation and understanding, and medical diagnosis.}
\label{fig:fig1}
\vspace{-0.5cm}
\end{figure}
Multimodal large language models (MLLMs) have witnessed rapid progress in recent years, driving remarkable advances across vision–language tasks spanning text recognition, image understanding and other multimodal scenarios~\citep{bai2025qwen2,wang2025internvl3,lu2025ovis2}. However, the MLLM landscape is highly heterogeneous: models vary substantially in architecture, training data, modality alignment, and computational efficiency~\citep{yin2024survey,wu2023multimodal}. We refer to this diversity as the \textbf{MLLM zoo}—a dynamic ecosystem of models with distinct strengths, limitations, and resource requirements.

This diversity, while fueling innovation, also reveals a critical limitation: \textbf{no single MLLM is uniformly superior across tasks.} In real-world deployments, tasks vary widely, ranging from lightweight OCR to complex multimodal reasoning~\citep{yue2024mmmu,liu2024ocrbench,lu2023mathvista}. As a result, using a single MLLM for all queries either over-allocates compute on easy instances or sacrifices accuracy on difficult ones.

\begin{figure}[ht]
\centering
\includegraphics[width=\linewidth]{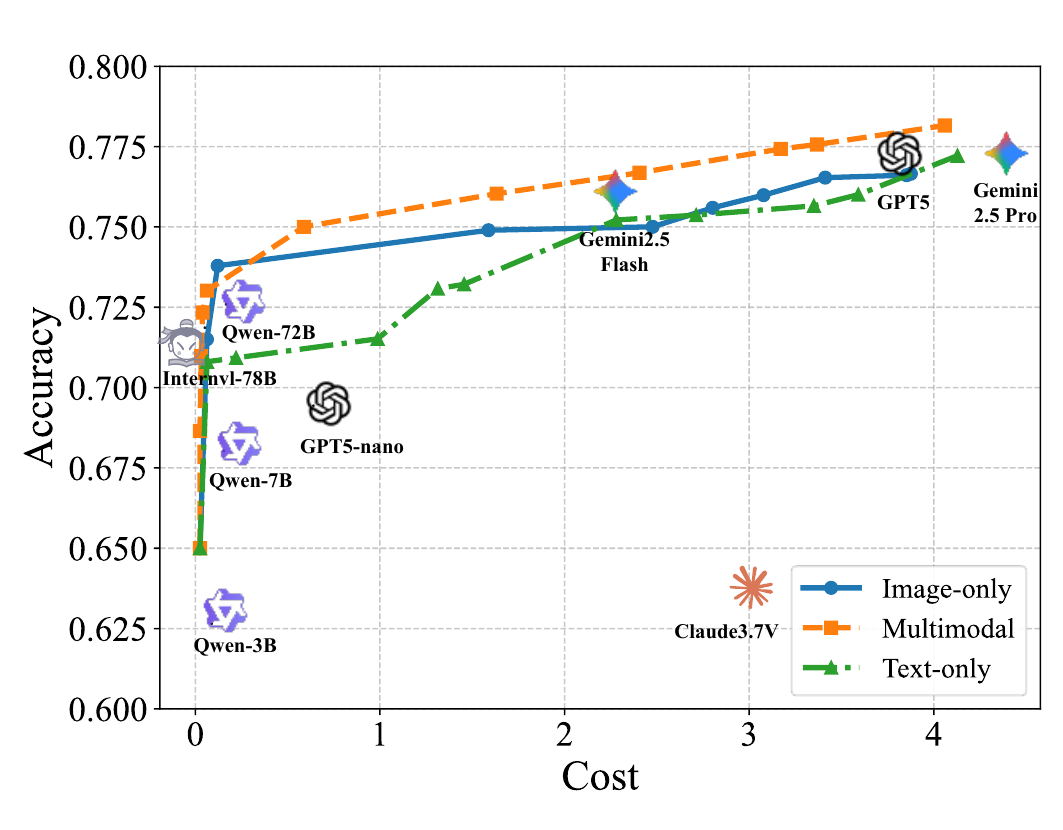}
\caption{Routing comparison across different modality signals, where Qwen denotes the Qwen2.5-VL series and InternVL denotes the InternVL3 series.}
\label{fig:fig2}
\vspace{-0.5cm}
\end{figure}

A natural remedy is routing, which has demonstrated its effectiveness for LLMs~\citep{chen2024routerdc,chen2023frugalgpt,feng2024graphrouter}. Routing selects different models per query to balance efficiency and accuracy. To examine whether this idea transfers to MLLMs, we simply applied a \textbf{text-only} routing strategy to multimodal tasks. Surprisingly, as shown in \Cref{fig:fig2}, text-only routing can \textbf{outperform} Qwen2.5-VL-7B and GPT5-nano at equal cost, indicating the strong potential of model routing. Consistent with intuition, however, it does not surpass several stronger models such as Gemini and GPT5 under the same cost; this suggests that text-only information is insufficient and \textbf{highlights the importance of incorporating multimodal signals} to further enhance routing performance.



Building on these findings, we introduce the concept of \textit{multimodal LLM routing}. As illustrated in \Cref{fig:fig1}, a user provides multimodal inputs along with a compute budget to the routing system, which selects an appropriate model from the MLLM zoo to generate the response and returns the output to the user.
Compared with routing strategies under text-only LLMs, this setting presents several unique challenges: (1) \textbf{Modality Gap}: MLLMs must integrate heterogeneous inputs across different modalities, introducing alignment and representation challenges absent in text-only scenarios~\citep{liang2022mind}; (2) \textbf{Model Heterogeneity}: the MLLM zoo comprises models with diverse architectures, training data, modality alignments, and computational characteristics, which complicates routing decisions; (3)  \textbf{Benchmarking Complexity}: routing for MLLMs couples multi-dimensional costs with modality-specific behavior, making it difficult to define unified cost models and candidate sets. \emph{This complexity has so far prevented the emergence of a standardized, modality-aware benchmark for MLLM routing, even though analogous benchmarks already exist for text-only LLMs}~\citep{lu2025routerarena,huang2025routereval,hu2024routerbench}.

These differences \textbf{motivate} the need for a controlled benchmark that isolates the multimodal routing problem and enables rigorous, directly comparable evaluations. We present \textbf{MMR-Bench}, designed to answer two key questions: (1) Are unimodal signals (text-only or image-only) sufficient, or are genuinely multimodal cues required? (2) Under fixed compute budgets, how much performance does multimodal routing actually deliver in practice?

Accordingly, MMR-Bench provides a unified framework for evaluating routing strategies across diverse MLLMs and tasks. It offers a controlled environment with standardized cost models, adjustable compute budgets, and modality-aware inputs; integrates a broad suite of vision–language benchmarks; and includes strong single-model baselines, oracle upper bounds, and representative routing methods. Together, these components enable consistent, reproducible comparisons and establish a solid foundation for studying adaptive multimodal model selection and efficient MLLM deployment.

Using MMR-Bench, we report three consistent findings:  
\begin{itemize}[leftmargin=*]
\item \textbf{Multimodal signals matter.} Unimodal routers are often miscalibrated on text-in-image inputs and in low-resolution or cluttered scenes, whereas fusing image and text modalities closes this gap at comparable cost.  
\item \textbf{Efficiency under budgets.} In certain scenarios, with only about 33\% of the cost of the strongest single model, our multimodal routing matches or even surpasses its accuracy, achieving a strictly better cost–accuracy trade-off.  
\item \textbf{Robust generalization.} A routing policy trained on a subset of MLLMs and tasks generalizes to unseen data from different distributions without additional tuning, and can also transfer effectively to text-only LLM routing.
\end{itemize}

\section{Related work}
\label{sec:related}

\subsection{Multimodal Large Language Models.}
Modern MLLMs couple language backbones with visual encoders through lightweight integration and instruction tuning, achieving broad vision–language competence~\citep{liu2023visual,li2024llava}. The ecosystem, however, spans markedly different cost profiles. \textbf{Commercial frontier models} typically provide stronger multi-step reasoning, long-context handling, and more robust perception~\citep{achiam2023gpt,team2023gemini}, while \textbf{open-weight models} are generally cheaper to deploy and remain competitive on routine recognition and short-form understanding tasks~\citep{wang2025internvl3,bai2025qwen2,lu2025ovis2}. This heterogeneity makes any single model suboptimal for all inputs and budgets. These observations motivate the idea of \emph{routing}, which decides at inference time which model to invoke for a given instance under explicit compute constraints~\citep{shnitzer2023large,zhang2025capability}. Our focus is on the selection problem rather than proposing new architectures, quantifying when lightweight models suffice and when heavier models are warranted.

\subsection{Routing and Model Selection for MLLMs.}
In text-only settings, model routing reduces cost without sacrificing quality by learning to dispatch queries among heterogeneous LLMs~\citep{song2025irt,yang2025quality}, and recent benchmarks~\citep{hu2024routerbench,huang2025routereval} have standardized evaluation across routers and cost–accuracy trade-offs. Extending these ideas to multimodal settings introduces new challenges. Routing must exploit cross-modal signals and handle architecture-level heterogeneity, making both confidence estimation and cost modeling more complex. Early multimodal approaches embed routing within an MLLM via Mixture-of-Experts architectures~\citep{lin2024moe} or retrofit existing MLLMs with dynamic expert paths~\citep{wu2024routing,aggarwal2024automix,vsakota2024fly}, improving efficiency and accuracy within a single model family but not addressing inter-model selection across the broader MLLM zoo. Our benchmark directly targets this gap by learning when to use which MLLM under a given budget.

\section{Preliminaries}

\subsection{Problem definition}

Let $M$ denote the number of modalities (e.g., text, image, audio, video).  
A \emph{multi-modal large language model} (MLLM) is a mapping
\[
f: \mathcal{X}^{1}\times\mathcal{X}^{2}\times\cdots\times\mathcal{X}^{M}\to\mathcal{Y},
\]
where $\mathcal{X}^r$ is the input space of modality $r$ and $\mathcal{Y}$ is the shared output space.

We focus on a routing problem in the \textbf{two-modality case} (text and image). For each incoming multimodal input, the goal is to select one model from a candidate zoo so as to achieve a desirable trade-off between model utility and cost. In practice, not all modalities are always available; to capture this, we introduce a \emph{modality availability vector}
\[
\mathbf{m}_i = (m_i^{\text{text}},\, m_i^{\text{img}}), \qquad m_i^{r} \in \{0,1\},
\]
where $m_i^{r}=1$ indicates modality $r$ is present and $m_i^{r}=0$ indicates it is missing. For example, $\mathbf{m}_i=(1,0)$ denotes a text-only sample, while $\mathbf{m}_i=(1,1)$ denotes a paired text--image input.

\begin{definition}[MLLM routing]
Define a support set
\begin{equation}\label{eq:support-set}
\mathcal{S} = \{ (x_i, \mathbf{m}_i, \mathbf{u}_i, \mathbf{c}_i) \}_{i=1}^n,
\end{equation}
where $x_i = (x_i^{\text{text}}, x_i^{\text{img}})$,  
$\mathbf{u}_i = (u_{i,1}, \dots, u_{i,K})^\top$ contains the utility of each candidate model on sample $i$,  
and $\mathbf{c}_i = (c_{i,1}, \dots, c_{i,K})^\top$ contains their corresponding costs.

For each sample the router selects the model index
\begin{equation}\label{eq:routing}
j_i^\star = \argmax_{j \in \{1, \dots, K\}} S_{i,j},
\end{equation}
using the overall score
\begin{equation}\label{eq:score}
S_{i,j} = u_{i,j} - \lambda\, c_{i,j},
\end{equation}
where $\lambda \ge 0$ is a tunable parameter that controls the trade-off between performance and cost.

\end{definition}

During inference, the router bases its decision only on the modalities that are available for a given input, as indicated by $\mathbf{m}_i$.  
Depending on the availability pattern, we can instantiate different baseline routers, such as \textit{text-only, image-only, or multimodal}, by leveraging corresponding feature extractors and fusion strategies.  
The modality availability vector $\mathbf{m}_i$ thus determines which router family is active and what information can be used in routing, and differences in routing performance directly reflect the contribution of each modality to the decision process.

\subsection{Motivation}
We conduct an empirical study using pre-computed utilities $\{u_{i,j}\}$ and costs $\{c_{i,j}\}$ from a heterogeneous zoo of MLLMs (\Cref{sec:construction}). The study addresses two questions: (i) whether a \emph{text-only} router is effective when inputs are multimodal, and (ii) whether unimodal observables (text-only or image-only) are sufficient, or whether genuinely multimodal cues are required.

\noindent \textbf{Setting.}
We consider a fixed set of candidate models $\{f_j\}_{j=1}^K$, along with their utilities $\{u_{i,j}\}$ and costs $\{c_{i,j}\}$ for each query. The only difference between routers lies in the observable features $\phi(x_i)$ they use: (a) \emph{text-only} features $\phi_{\text{text}}(x_i)$; (b) \emph{image-only} features $\phi_{\text{img}}(x_i)$; and (c) \emph{multimodal} features $\phi_{\text{mm}}(x_i)$ that jointly encode text and image signals. Each router defines a policy $\pi(x_i)\!\in\!\{1,\dots,K\}$ and is evaluated with the score defined by \Cref{eq:score}, where $\lambda\!\ge\!0$ is varied to trace the cost–accuracy trade-off.

\noindent \textbf{Protocol.}
To avoid confounding effects from powerful learned routers, we employ a lightweight procedure shared across all routing families.
Given a trade-off parameter $\lambda \ge 0$ and a specified number of clusters $k$, the protocol is as follows:
(i) compute features $\phi(x_i)$ on the training and validation split;
(ii) apply $k$-means to the training features to obtain clusters $\{h\}_{h=1}^k$;
(iii) for each cluster $h$, select a single model by maximizing the average validation score:
\begin{equation}\label{eq:cluster-assign}
  j_h^\star(\lambda)
  = \argmax_{j \in \{1,\dots,K\}}
    \frac{1}{|\mathcal{S}_{\text{val}}(h)|}
    \sum_{i \in \mathcal{S}_{\text{val}}(h)}
    S_{i,j}(\lambda),
\end{equation}
where $\mathcal{S}_{\text{val}}(h)$ denotes validation samples assigned to cluster $h$, and $S_{i,j}(\lambda)$ is defined in \Cref{eq:score}. 
At test time, each sample $x$ is mapped to its nearest cluster $h(x)$ and routed to the corresponding selected model $j^\star_{h(x)}$.
By sweeping over the number of clusters $k$ and the cost weight $\lambda$, we generate a family of routing policies spanning different points on the cost–accuracy spectrum. Their frontiers (\Cref{fig:fig2}) allow comparison under matched-cost and matched-accuracy settings.

\begin{mdframed}[backgroundcolor=gray!10, linecolor=black, linewidth=1pt]
\textbf{\textit{Q1:}} Can text-only routing methods be used in multimodal scenarios?
\end{mdframed}

Yes, but only partially. Across mixed multimodal workloads, the text-only router improves over several fixed models such as Qwen2.5-VL-7B and GPT-5-nano at comparable normalized cost as illustrated in Fig.~\ref{fig:fig2}, which shows that instance-wise selection can help even without visual features. However, the improvements are uneven. The router consistently underperforms on subsets where visual evidence determines difficulty, including text-in-image queries, dense OCR, charts, spatial reasoning, and low-resolution crops. Error analysis indicates that these failures arise from incorrect assignments: the router tends to overestimate the adequacy of cheaper models when textual cues provide an unreliable proxy for visual complexity. Overall, text-only routing fails to achieve optimal performance on vision-governed cases. \textit{Detailed case are provided in the supplementary material.}

\begin{mdframed}[backgroundcolor=gray!10, linecolor=black, linewidth=1pt]
\textbf{\textit{Q2:}} Are unimodal signals sufficient, or are multimodal cues required?
\end{mdframed}

Multimodal cues are essential for strong routing. \Cref{fig:fig2} compares routing frontiers under identical model sets and cost accounting. The \emph{multimodal} router consistently outperforms both \emph{text-only} and \emph{image-only} variants: at fixed cost it achieves higher accuracy, and at fixed accuracy it operates at lower cost. Ablations over the number of clusters $k$ and the cost weight $\lambda$ show that this advantage is stable, and the gap widens in settings where visual clutter or linguistic complexity dominates difficulty. Category-level evaluations show the same trend. Text-only routing falls behind on vision-driven inputs, image-only routing falls behind on language-driven inputs, and multimodal features reduce incorrect assignment in both regimes.

\noindent \textbf{Takeaways.}
(1) Routing over heterogeneous MLLMs improves the cost–accuracy trade-off even with simple, unsupervised policies. (2) Unimodal observables are insufficient to realize the full gains; access to \emph{joint} text–image cues is necessary to approach the empirical frontier on mixed multimodal workloads.

\section{MMR-Bench: Benchmark Construction}
\label{sec:construction}

We introduce \textbf{MMR-Bench}, a budget-aware benchmark for routing across the heterogeneous MLLM zoo of candidate models.
MMR-Bench is designed as an \emph{offline environment}: for each multimodal query and each candidate model, we provide the raw model output, a task-specific utility score, and a normalized inference cost derived under a unified pricing scheme.
This outcome table (over \textbf{100k} pairs of instances and models across multiple candidates and tasks), together with frozen splits and deterministic scoring scripts, enables systematic and reproducible evaluation of cost-aware routing strategies without re-running any MLLM, as summarized in \Cref{tab:bench-compare}.

\begin{table}[t]
\centering
\caption{Comparison across routing benchmarks along four dimensions: supported modality, reported evaluation metrics, and the availability (or count) of router baselines. MMR-Bench reports three cost-aware metrics in an offline setting.}
\label{tab:bench-compare}
\small
\renewcommand{\arraystretch}{1.15}
\setlength{\tabcolsep}{5pt}
\begin{tabular}{lcccc}
\toprule
\textbf{Benchmark} & \textbf{Text} & \textbf{Image} & \textbf{Metrics} & \textbf{Baselines} \\
\midrule
RouterBench~\cite{hu2024routerbench} & \cmark\ & \xmark\  & 1  &  5 \\ 
RouterEval~\cite{huang2025routereval}  & \cmark\ & \xmark\  & 1  &  8 \\ 
EmbedLLM~\cite{zhuang2024embedllm}    & \cmark\ & \xmark\  & 1 &  4 \\ 
MixInstruct~\cite{jiang2023llm} & \cmark\ & \xmark\  & 3 & 6 \\ \hline
\addlinespace[3pt]
\textbf{MMR-Bench} & \cmark\ & \cmark\ &  3 &  10 \\
\bottomrule
\end{tabular}
\end{table}

\subsection{Design Principles}
\label{sec:design}

MMR-Bench is built specifically for routing across a heterogeneous MLLM zoo, rather than as a generic vision–language leaderboard. Its construction follows three principles:

\begin{itemize}
    \item \textbf{Extensive coverage.}
    The dataset selection covers most typical multimodal scenarios and difficulty levels, with varied resolutions and scene complexity.

    \item \textbf{Representative models.}
    The candidate pool includes widely used MLLMs with diverse architectures, capacities, and training regimes, ensuring practical relevance.

    \item \textbf{Extensibility.}
    The benchmark is offline and modular; new models or datasets can be added without re-running existing ones, while frozen splits and deterministic scoring preserve comparability.
\end{itemize}

\subsection{Datasets and Model Pool}
\label{sec:datasets-experts}

\begin{figure}[t]
\centering
\includegraphics[width=0.8\linewidth]{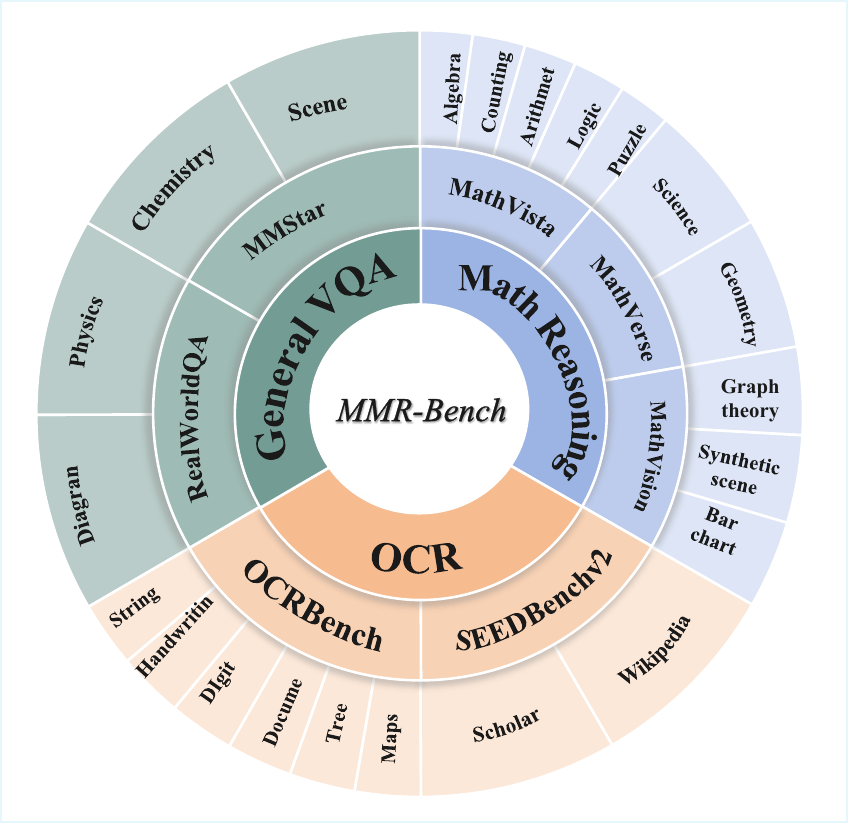}
\caption{Composition of \textbf{MMR-Bench} across scenarios and datasets.}
\label{fig:mmr-pie}
\end{figure}

We focus on three routing-relevant multimodal scenarios and instantiate them using established benchmarks to cover diverse difficulty and varying reliance on modalities:
\begin{enumerate}[label=(\arabic*), leftmargin=*]
\item \textbf{Document-centric OCR and understanding:} using OCRBench~\cite{liu2024ocrbench} and SEED-Bench v2 Plus~\cite{li2023seed}, which contain scanned pages, dense layouts, and long-form documents that stress OCR quality, layout parsing, and long-context reasoning.
\item \textbf{General VQA and grounding:} using MMStar~\cite{chen2024we} and RealWorldQA~\cite{malinowski2014multi}, covering natural images, screenshots, charts, and real-world scenes with queries ranging from recognition to fine-grained grounding and robustness tests.
\item \textbf{Multimodal math and diagram reasoning:} using MathVerse~\cite{zhang2024mathverse}, MathVista~\cite{lu2023mathvista}, and MathVision~\cite{wang2024measuring}, which combine text, equations, and diagrams and require compositional reasoning, symbolic manipulation, and precise visual understanding.
\end{enumerate}
These datasets jointly ensure that no single MLLM is uniformly optimal and that instance-wise, budget-aware routing offers measurable headroom over single-model and unimodal baselines, with the overall benchmark composition shown in Fig.~\ref{fig:mmr-pie}.

\noindent \textbf{Model zoo.}
The model set $\mathcal{F} = \{f_j\}_{j=1}^K$ includes both closed-source and open-weight MLLMs with heterogeneous architectures, capacities, and cost profiles:
\begin{itemize}
\item \textbf{Commercial MLLMs}, including GPT-5 series~\cite{achiam2023gpt}, Gemini 2.5 series~\cite{team2023gemini}, and Claude models, which offer strong general-purpose multimodal reasoning at higher monetary cost.
\item \textbf{Open-weight MLLMs}, spanning 3B–72B parameters and including Gemma~\cite{team2025gemma}, InternVL3~\cite{wang2025internvl3}, and Qwen2.5-VL families~\cite{bai2025qwen2}, which are cheaper to deploy yet exhibit diverse OCR, grounding, and reasoning behavior.
\end{itemize}

All models share fixed prompts and decoding settings within each scenario.
For every instance–model pair, we provide a task-specific utility $u_{i,j} \in [0,1]$ and a normalized inference cost $c_{i,j} \ge 0$, forming the offline outcome table used throughout our routing evaluation.

In total, MMR-Bench comprises 11{,}000 instances covering 10 models, 8 datasets, and 3 scenarios.

\subsection{Cost-aware Routing Evaluation Protocol}
\label{sec:protocol}

MMR-Bench evaluates routing policies entirely in an \emph{offline} fashion using the precomputed instance–model outcomes from \Cref{eq:support-set}. Following that formulation, we use the support set
\begin{equation}
\mathcal{S}=\big\{\,((x_i^{\text{text}},x_i^{\text{img}}),\mathbf{u}_i,\mathbf{c}_i)\,\big\}_{i=1}^n.
\end{equation}

\noindent \textbf{Routing behavior.}
For an input $x_i$ (optionally with a budget signal $b_i$), a routing policy $R_\theta$ outputs a distribution over models
\begin{equation}
R_\theta(x_i, b_i) \in \Delta^{K-1}, \quad
\pi_\theta(j \mid x_i, b_i) = [R_\theta(x_i, b_i)]_j.
\end{equation}

The per-instance expected utility and cost are obtained by weighting the precomputed $\{u_{i,j},c_{i,j}\}$ with the router’s distribution $\pi_\theta(\cdot \mid x_i,b_i)$; dataset-level means summarize each method as $\mathrm{Perf}(R_\theta)$ and $\mathrm{Cost}(R_\theta)$. Varying operating points produces $(\mathrm{Cost},\mathrm{Perf})$ pairs whose Pareto upper envelope defines the performance–cost curve $p(c)$, evaluated over the shared range $[c_{\min},c_{\max}]$ set by always choosing the cheapest and the most expensive single-model baselines.

\noindent \textbf{Routing constraints.}
Routers may use only inference-time information: (i) the query $(x_i^{\text{text}},x_i^{\text{img}})$ and optional budget $b_i$; (ii) benchmark lightweight features (e.g., embeddings, token counts, scenario tags); (iii) static model metadata (e.g., normalized cost, context length, supported modalities). They may not access ground-truth labels or $\{u_{i,j}\}$ on the evaluation split, nor issue multiple adaptive model calls per instance, since MMR-Bench fixes one outcome per instance–model pair. Oracle selectors (e.g., $\arg\max_j u_{i,j}$) are reported only as analysis upper bounds.

\noindent \textbf{Metrics.}
All utilities are normalized to $[0,1]$ to enable aggregation across tasks and models. We summarize each routing method with three complementary metrics derived from $p(c)$.

\noindent\textbf{(1) nAUC (Normalized AUC).}
The area under the performance–cost curve, normalized by the range length across the evaluated cost interval:
\begin{equation}
\mathrm{nAUC}
= \frac{1}{c_{\max} - c_{\min}}
  \int_{c_{\min}}^{c_{\max}} p(c)\, \mathrm{d}c.
\end{equation}

where $p(c)$ is the performance at cost $c$.

\noindent\textbf{(2) Peak score ($P_s$)~\cite{jitkrittum2025universal,zhuang2024embedllm}.}
The maximal performance observed on the curve:
\begin{equation}
P_s \;=\; \max_{c} \, p(c).
\end{equation}

\noindent\textbf{(3) Quality–Neutral Cost (QNC)~\cite{jitkrittum2025universal}.}
Let $(p_{\text{best}}, c_{\text{best}})$ denote the performance and cost of the most accurate single model in the pool. QNC is the minimum relative cost required for the router to reach the same performance:
\begin{equation}
\mathrm{QNC}
= \frac{1}{c_{\text{best}}}
  \min \{\, c \mid p(c) \ge p_{\text{best}} \,\}.
\end{equation}

If $p(c) < p_{\text{best}}$ for all $c \in [c_{\min}, c_{\max}]$, we report $\mathrm{QNC}=+\infty$ and mark the target quality as not reached.

\begin{table*}[t]\small
\caption{Performance of different routing methods across datasets and settings. \textbf{Bold numbers} indicate the best performance (excluding Oracle), and \underline{underlined numbers} indicate the second-best (excluding Oracle).}
\label{tab:main_results1}
\centering
\renewcommand{\arraystretch}{0.9}
\begin{tabular}{lcccccccc}
\toprule
Scenario & \multicolumn{4}{c}{\textbf{OCR}} & \multicolumn{4}{c}{\textbf{General VQA}} \\
\cmidrule(lr){1-1} \cmidrule(lr){2-5} \cmidrule(lr){6-9}
Dataset & \multicolumn{2}{c}{OCRBench} & \multicolumn{2}{c}{SeedBenchV2Plus} & \multicolumn{2}{c}{MMStar} & \multicolumn{2}{c}{RealWorldQA} \\
\cmidrule(lr){1-1} \cmidrule(lr){2-3} \cmidrule(lr){4-5} \cmidrule(lr){6-7} \cmidrule(lr){8-9}
\diagbox{Method}{Metric}
  & nAUC $(\uparrow)$ & $p_s (\uparrow)$
  & nAUC $(\uparrow)$ & $p_s (\uparrow)$
  & nAUC $(\uparrow)$ & $p_s (\uparrow)$
  & nAUC $(\uparrow)$ & $p_s (\uparrow)$ \\
\midrule
Random    & --       & 0.8075 & --       & 0.7102 & --       & 0.6942 & --       & 0.7108 \\
LinearMFRouter   & 0.8789   & 0.9088 & 0.7376   & 0.7458 & 0.7496   & 0.7683 & 0.8036   & \underline{0.8252} \\
MLPMFRouter      & \underline{0.8952} & 0.9088 & \textbf{0.7443} & \textbf{0.7508} & \textbf{0.7615} & \textbf{0.7817} & \textbf{0.8097} & \textbf{0.8268} \\
KNNRouter        & 0.8913   & \textbf{0.9150} & 0.7375   & 0.7425 & 0.7560   & \underline{0.7767} & 0.8009   & \underline{0.8252} \\
KMeansRouter     & \textbf{0.9126} & \underline{0.9138} & \underline{0.7382} & \underline{0.7475} & \underline{0.7564} & 0.7733 & \underline{0.8072} & 0.8235 \\
\midrule
Oracle           & 0.9663   & 0.9825 & 0.8854   & 0.8913 & 0.9563   & 0.9650 & 0.9773   & 0.9853 \\
Best Single Model     & 1.0000 & 0.9050 & 1.0000   & 0.7502 & 1.0000 & 0.7716 & 1.0000   & 0.8235 \\
\bottomrule
\\
\toprule
Scenario & \multicolumn{6}{c}{\textbf{Math reasoning}} & \multicolumn{2}{c}{\textbf{Avg}} \\
\cmidrule(lr){1-1} \cmidrule(lr){2-7} \cmidrule(lr){8-9}
Dataset & \multicolumn{2}{c}{MathVista} & \multicolumn{2}{c}{MathVerse} & \multicolumn{2}{c}{MathVision} & \multicolumn{2}{c}{Full Dataset} \\
\cmidrule(lr){1-1} \cmidrule(lr){2-3} \cmidrule(lr){4-5} \cmidrule(lr){6-7} \cmidrule(lr){8-9}
\diagbox{Method}{Metric}
  & nAUC $(\uparrow)$ & $p_s (\uparrow)$
  & nAUC $(\uparrow)$ & $p_s (\uparrow)$
  & nAUC $(\uparrow)$ & $p_s (\uparrow)$
  & nAUC $(\uparrow)$ & $p_s (\uparrow)$ \\
\midrule
Random    & --       & 0.6688 & --       & 0.4800 & --       & 0.3927 & --       & 0.6135 \\
LinearMFRouter   & 0.7724   & 0.8038 & \underline{0.6837} & \underline{0.7258} & 0.5565   & 0.6826 & \textbf{0.7042} & \textbf{0.7533} \\
MLPMFRouter      & 0.7751   & \underline{0.8038} & 0.6784   & \textbf{0.7337} & 0.5494   & \textbf{0.6937} & 0.6913   & 0.7494 \\
KNNRouter        & \textbf{0.7807} & \textbf{0.8075} & \textbf{0.6881} & 0.7194 & \underline{0.5610} & 0.6867 & \underline{0.6950} & 0.7457 \\
KMeansRouter     & \underline{0.7763} & \underline{0.8038} & 0.6835   & 0.7194 & \textbf{0.5639} & \underline{0.6933} & 0.6885   & \underline{0.7496} \\
\midrule
Oracle           & 0.9310   & 0.9475 & 0.8806   & 0.8954 & 0.8031   & 0.8741 & 0.8897   & 0.9188 \\
Best Single Model     & 1.0000 & 0.8037 & 1.0000   & 0.7194 & 1.0000 & 0.6932 & 1.0000   & 0.7412 \\
\bottomrule
\end{tabular}
\end{table*}

\section{Experiments}
\label{sec:experiments}

\subsection{Experimental Setup}
\label{sec:exp-setup}

\noindent \textbf{Datasets and splits.}
We use the three scenarios and eight datasets in Sec.~\ref{sec:datasets-experts} with frozen 2:8 train/test splits. Routers are trained on the training split, test-time evaluation is purely offline by applying the learned router to benchmark features and indexing into the fixed $\{u_{i,j},c_{i,j}\}$ (Sec.~\ref{sec:protocol}).

\noindent \textbf{Models and cost.}
We route over the $K{=}10$ candidate MLLMs in Sec.~\ref{sec:datasets-experts}. All per-instance utilities $u_{i,j}\!\in[0,1]$ and normalized monetary costs $c_{i,j}\!\in\mathbb{R}_+$ are precomputed under a unified cost scheme and shared by all methods. The shared cost range $[c_{\min},c_{\max}]$ is defined by always selecting the cheapest versus the most expensive single-model.

\noindent \textbf{Metrics and aggregation.}
Following Sec.~\ref{sec:protocol}, we report the performance--cost curve $p(c)$ and summarize each method in the main text with nAUC and $P_s$ (peak score). The QNC metric is reported in the appendix for completeness. Metrics are computed per dataset and then macro-averaged within each scenario and across all scenarios. Unless otherwise noted, results for stochastic routers are averaged over multiple runs.

\subsection{Baselines}
\label{sec:exp-baselines}

\noindent \textbf{Feature fusion.}
We first extract frozen embeddings offline: images are encoded with a ViT-based encoder and text with a pretrained language encoder. We then construct a per-instance feature $z_i$ from these embeddings as follows:

\begin{itemize}[leftmargin=*]
  \item \textbf{Equal-weight mean.} Align the text and image embedding dimensions; if a modality is missing, fill with a zero vector. Fuse by averaging the two embeddings with equal weight to obtain a single joint representation.

  \item \textbf{Adaptive weights.} First estimate a confidence for each modality using two cues—cosine similarity to the batch mean and a norm-based sigmoid score—then turn these confidences into softmax weights. Build three components: a weighted sum (captures overall signal), an elementwise product (captures cross-modal agreement), and an absolute difference (captures mismatch). Linearly combine these components and apply $\ell_2$ normalization to form the final fused feature.
\end{itemize}

We analyze fusion choices later in Sec.~\ref{ana}; implementation details are in \textit{supplementary materials}.

\noindent \textbf{Routing methods.}
All methods take \(z_i\) (and optionally a budget \(b_i\)) and output a routing decision over the \(K\) candidate models:
\begin{itemize}[leftmargin=*]
    \item \textbf{Random (lower bound)}: For any query, a model is selected uniformly at random. This provides the theoretical lower bound.

    \item \textbf{Oracle (upper bound)}: Assumes that the router has access to each model’s performance and cost without performing inference, enabling the optimal choice. This provides the theoretical upper bound.

    \item \textbf{LinearRouter}: Formulates routing as a multi-output linear regression problem that predicts the utility and cost of each model. Variants using matrix factorization (MF) are also included.

    \item \textbf{MLPRouter}: Trains a shallow MLP regressor to predict each model’s utility and cost. Variants incorporate MF to improve representation capacity.
    
    \item \textbf{KNNRouter}: Applies $k$-nearest neighbors in feature space, scoring each candidate model by averaging the utility and cost of its nearest neighbors.
    
    \item \textbf{KMeansRouter}: Runs $k$-means clustering over training features to obtain cluster assignments. Each model’s performance across clusters serves as its capability code, which is used to make routing decisions at test time.
\end{itemize}

\noindent \textbf{Training and evaluation.}
Learned routers are trained on the training split, test-time evaluation is purely offline by indexing into the fixed \(\{u_{i,j},c_{i,j}\}\) under the normalized cost scheme (Sec.~\ref{sec:protocol}). All methods use the same \(z_i\) for a controlled comparison.

\subsection{Overall Results on MMR-Bench}
\label{sec:overall-results}

\noindent \textbf{Comparative Analysis.}
\Cref{tab:main_results1} shows that matrix-factorization (MF) based routers offer the most reliable performance across heterogeneous workloads. On the full-dataset average, LinearMFRouter attains the highest nAUC and peak score (0.7042 and 0.7533), with MLPMFRouter close behind on $P_s$ (0.7494) and competitive on nAUC (0.6913). Instance-based methods yield localized wins: KNNRouter achieves the best $P_s$ on \emph{OCRBench} (0.9150) and the top nAUC and $P_s$ on \emph{MathVista} (0.7807 and 0.8075), while KMeansRouter leads nAUC on \emph{OCRBench} (0.9126) and on \emph{MathVision} (0.5639, second-best $P_s$ 0.6933). MF routers are consistently superior on general VQA (\emph{MMStar} and \emph{RealWorldQA}), indicating better cross-category robustness. In short, modeling per-model outcomes in a low-rank space yields smoother generalization, whereas instance-based or clustering heuristics can peak on specific distributions but are less stable in aggregate.

\noindent \textbf{Cost efficiency and stability.}
The results point to a general conclusion: routers that \emph{explicitly model per–model outcomes in a low-rank space} produce smoother, budget-robust frontiers. Their higher macro nAUC and competitive $P_s$ indicate sustained quality across budgets rather than spikes at isolated operating points, suggesting they capture transferable, modality-aware difficulty signals rather than dataset-specific quirks. In contrast, purely instance-based or clustering dispatchers can peak on narrow distributions but are brittle when budgets or domains shift. \textit{Practical takeaway:} under variable or uncertain budgets, MF-style routers are the safer default; instance-based variants are best reserved for stable, well-characterized workloads.

\begin{figure}[t]
    \centering
    \includegraphics[width=0.95\linewidth]{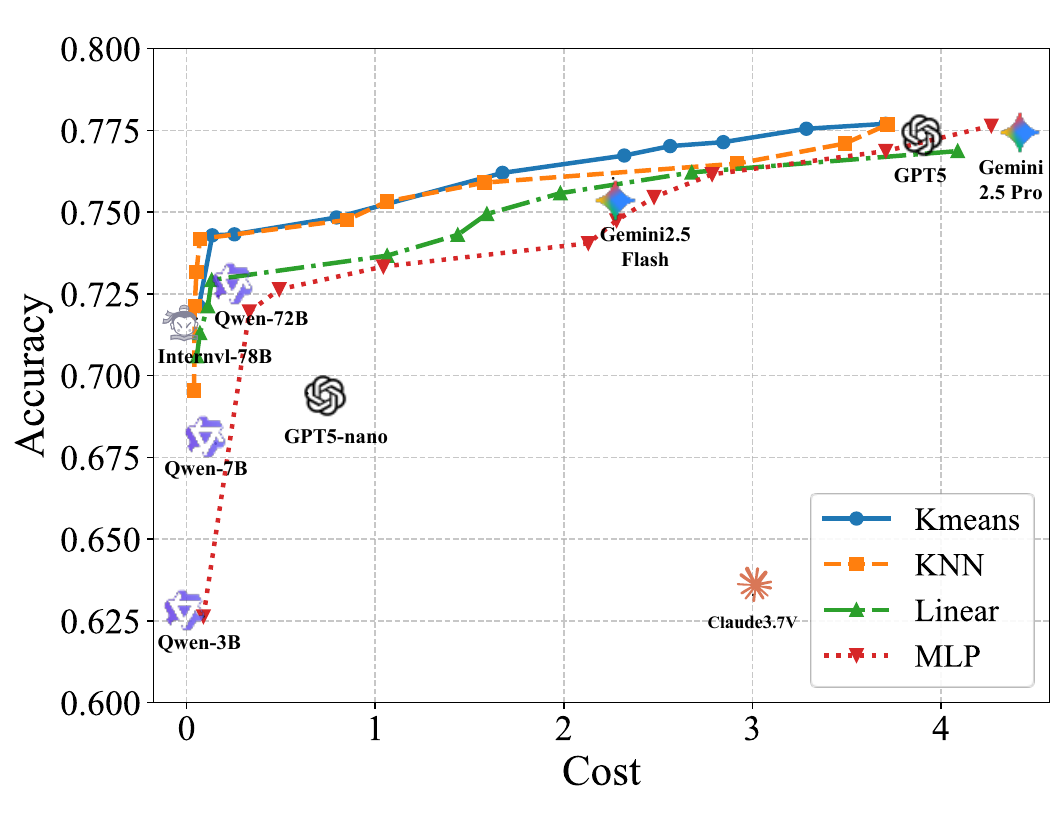}
    \caption{
        \textbf{Cost-accuracy Pareto frontiers on MMR-Bench}. Lines of different styles represent different routing strategies, while single-model positions are indicated by their logos. For readability, the x-axis is log-scaled to emphasize the low-cost region. Compared with any fixed model, routing shifts the frontier upward and leftward, indicating higher accuracy at the same cost and lower cost at the same accuracy.
    }
    \label{fig:pareto-front}
\end{figure}

\noindent \textbf{Pareto-front comparison.}
To make the budget--quality trade-off visually clear, we plot accuracy against normalized cost and report the Pareto envelope of each router, with single-model baselines shown as isolated points (\Cref{fig:pareto-front}). We rescale the $x$-axis to magnify the low-cost region where most practical operating points lie. The resulting frontiers show that instance-wise routing consistently dominates fixed-model choices: with only \textbf{33$\%$} of the strongest single model’s cost, our router already matches its accuracy, and at higher budgets it \emph{surpasses} that model, yielding a uniformly stronger cost--accuracy trade-off. This demonstrates that adaptive selection over a heterogeneous MLLM pool is effective in practice, especially under tight or variable budget constraints.

\section{Analyses}
\label{ana}

Building on the overall routing results in Sec.~\ref{sec:overall-results}, this section briefly analyzes the sources of routing gains, focusing on modality gaps revealed by adaptive fusion, robustness to intra-scenario distribution shift, and cross-modal transfer from multimodal to unimodal tasks.

\begin{mdframed}[backgroundcolor=gray!10, linecolor=black, linewidth=1pt]
(1) Adaptive fusion reveals a clear modality gap between image and text signals.
\end{mdframed}

\vspace{-2pt}  
\Cref{tab:fusion-deltas} contrasts adaptive fusion with equal-weight averaging under fixed router families.
Adaptive yields the largest lift for \textbf{KMeans} (\(\Delta\)nAUC \(=+0.3403\); QNC \(+\infty\!\rightarrow\!1.0585\)),
small but consistent gains for \textbf{MLP} (QNC \(+\infty\!\rightarrow\!\mathbf{0.9947}<1\)),
modest average improvements for \textbf{Linear} (\(\Delta\)nAUC \(=+0.0124\)) with a slightly higher QNC (\(0.9055\!\rightarrow\!0.9701\)),
and mild regressions for \textbf{KNN} (\(\Delta\)nAUC \(=-0.0074\); QNC remains \(+\infty\)).

These outcomes provide direct evidence of a \emph{modality gap} in multimodal routing: the information carried by image and text is \emph{not} uniformly balanced across tasks or instances.
Equal-weight fusion imposes an implicit parity assumption between modalities; when one modality dominates or the modalities disagree, this assumption is violated, yielding miscalibrated scores and suboptimal model choices. Adaptive fusion explicitly \emph{reweights} modalities at the instance level and exposes \emph{agreement/mismatch} via multiplicative and difference interactions. Overall, these fusion results substantiate the modality-gap hypothesis: because modality importance varies across workloads, routers that sense and exploit this variation dominate naive equal averaging on the cost–performance frontier.

\begin{table}[t]
\centering
\caption{Adaptive minus equal fusion. Positive $\Delta$ is better for nAUC/$P_s$; lower QNC is better.}
\label{tab:fusion-deltas}
\small
\setlength{\tabcolsep}{6pt}
\begin{tabular}{lccc}
\toprule
\textbf{Method} & $\Delta$ nAUC & $\Delta P_s$ & \textbf{QNC change} \\
\midrule
KMeans & $+0.3403$ & $+0.1275$ & $+\infty \rightarrow 1.0585$ \\
MLP    & $+0.0033$ & $+0.0020$ & $+\infty \rightarrow 0.9947$ \\
Linear & $+0.0124$ & $+0.0036$ & $0.9055 \rightarrow 0.9701$ \\
KNN    & $-0.0074$ & $-0.0014$ & $+\infty \rightarrow +\infty$ \\
\bottomrule
\end{tabular}
\end{table}

\begin{mdframed}[backgroundcolor=gray!10, linecolor=black, linewidth=1pt]
(2) Routing policies can remain robust under intra-scenario distribution shifts.
\end{mdframed}
\vspace{-2pt}  
To test whether routing captures \emph{scenario-consistent}, modality-aware difficulty cues rather than overfitting to dataset-specific biases, we run a within-scenario cross-dataset evaluation: for each scenario, the router evaluated on a different dataset from the same scenario. As shown in \Cref{tab:transfer}, the cross-dataset router consistently achieves a higher peak score \(P_s\) than the best single model and remains close to the in-domain router.

\begin{table}[t]
\centering
\small
\caption{Within-scenario cross-dataset robustness measured by peak score \(P_s\) (\(\uparrow\) higher is better). 
Each column reports a scenario; rows compare the best single model on the target dataset \(B\) with the cross-dataset router \(A\!\to\!B\) (\textbf{Shift})}
\label{tab:transfer}
\setlength{\tabcolsep}{10pt}
\begin{tabular}{lccc}
\toprule
 Method & \textbf{OCR} & \textbf{VQA} & \textbf{Math} \\
\midrule
Best model
& 0.7062 & 0.7936 & 0.7592 \\
\textbf{Shift}
& \textbf{0.7234} & \textbf{0.8012} & \textbf{0.7914} \\
\bottomrule
\end{tabular}
\vspace{-0.5cm}
\end{table}

The cross-dataset results demonstrate robustness to within-scenario distribution shift: the router is not memorizing dataset quirks but capturing scenario-consistent, modality-aware difficulty signals. From a multimodal perspective, the transferable signals include instance-level modality salience (relative importance of image versus text), cross-modal agreement or mismatch, and structural cues such as OCR density, layout complexity, and question length. Our adaptive, interaction-augmented fusion exposes these cues through reweighting, product, and difference terms, which remain stable at the scenario level and thus support transfer. The small residual gap to the in-domain router is attributable to dataset-specific biases rather than a failure to generalize.

\begin{mdframed}[backgroundcolor=gray!10, linecolor=black, linewidth=1pt]
 (3) Multimodal routers can transfer across modalities and generalize to unimodal tasks.
\end{mdframed}
\vspace{-2pt}
Building on Sec.~\ref{ana}.(2), which established robustness to \emph{distribution} shift within scenarios, we now consider whether the router is also robust across \emph{modalities}. Concretely, we train the router on multimodal workloads using the same fusion features as in Sec.~\ref{sec:exp-baselines}, and then evaluate it on text-only benchmarks where only the textual observable is available at test time. At inference we \emph{freeze} the multimodal router and expose only the text channel by masking the image embedding in the fusion pipeline, yielding a zero-shot modality transfer setting. We compare against (i) the best single-model baseline on the text benchmark and (ii) our multimodal-trained router under text-only inference.

\begin{table}[t]
\centering
\small
\caption{Cross-modality transfer to \emph{text-only} benchmarks. The router is trained on \emph{multimodal} workloads and evaluated under text-only inference by masking the image channel. We report peak score $P_s$ (\(\uparrow\) higher is better). Rows compare (i) the best single-model baseline on each text benchmark and (ii) our multimodal-trained router evaluated in the text-only setting (MM$\!\rightarrow\!$Text).}
\label{tab:modality-transfer}
\setlength{\tabcolsep}{10pt}
\begin{tabular}{lccc}
\toprule
\textbf{Method} & \textbf{GSM8K} & \textbf{MMLU} & \textbf{ARC} \\
\midrule
Best single model & 94.5 & 91.2 & 65.7 \\
\textbf{MM$\!\rightarrow\!$Text (ours)} & \textbf{96.7} & \textbf{92.4} & \textbf{66.7} \\
\bottomrule
\end{tabular}
\vspace{-0.5cm}
\end{table}

As summarized in \Cref{tab:modality-transfer}, a router trained on multimodal data retains strong effectiveness when the image channel is absent: its $P_s$ consistently exceeds the best single model on GSM8K/MMLU/ARC. We attribute this transfer to three factors. \textit{(i) Modality-agnostic difficulty cues:} the router learns predictors tied to textual structure (e.g., problem form, token length, numeracy cues, chain-of-thought depth) that correlate with which model is needed, independent of visual content. \textit{(ii) Fusion-induced regularization:} training with interaction features (sum/product/difference) teaches the router to reason about \emph{agreement} and \emph{mismatch} across channels; when the image is masked at test time, these interactions degrade gracefully to text-only surrogates rather than collapsing. \textit{(iii) Cost-aware calibration:} optimizing for cost--performance encourages conservative expert selection under uncertainty, a behavior that remains valid when one modality is missing. Together, these effects yield a routing policy that is robust to modality drop and deployable on text-only workloads without retraining.

\section{Conclusion}
\label{sec:conclusion}
We present \textbf{MMR-Bench}, an offline, cost-aware benchmark for routing over a heterogeneous collection of multimodal large language models. By providing precomputed utilities and normalized costs for each instance–model pair, along with unified evaluation metrics, MMR-Bench enables controlled, reproducible comparison of routing policies without rerunning any model. Our experiments demonstrate that routing significantly improves the cost–accuracy trade-off over the best single model, sometimes matching or exceeding its accuracy at roughly one-third of the cost, with matrix-factorization-based routers achieving the most robust performance across heterogeneous workloads. Further analyses show that multimodal observables with adaptive fusion are critical for strong routing, closing the gap left by unimodal policies, and that the learned routers remain stable under intra-scenario distribution shifts and can transfer to text-only benchmarks. We hope that MMR-Bench will serve as a standardized testbed for future research on cost-aware multimodal routing and the practical deployment of MLLMs under realistic budget constraints.

{
    \small
    \bibliographystyle{ieeenat_fullname}
    \bibliography{main}
}


\clearpage 
\setcounter{page}{1} 
\setcounter{section}{0}
\setcounter{figure}{0}
\setcounter{table}{0}
\setcounter{equation}{0}

\renewcommand{\thesection}{\Alph{section}} 
\renewcommand{\thefigure}{S\arabic{figure}}
\renewcommand{\thetable}{S\arabic{table}}
\renewcommand{\theequation}{S\arabic{equation}}

\twocolumn[
  \begin{center}
    \Large\textbf{Supplementary Material for\\MMR-Bench: A Comprehensive Benchmark for Multimodal LLM Routing}
    \vspace{1.5em}
  \end{center}
]


\section{Overview}
This supplementary material provides additional details supporting the main paper, organized as follows:
\begin{itemize}
    \item \textbf{Section \ref{sec:supp_dataset}}: Detailed statistics and licensing information for the eight datasets used in MMR-Bench.
    \item \textbf{Section \ref{sec:supp_models}}: Specifications of the 10-model zoo and the cost normalization formulation.
    \item \textbf{Section \ref{sec:supp_imp}}: In-depth implementation details of the router architectures and the adaptive fusion mechanism (Eq. 6 in main paper).
    \item \textbf{Section \ref{sec:supp_exp}}: Complete experimental results, including the Quality-Neutral Cost (QNC) metric and full per-dataset breakdown.
    \item \textbf{Section \ref{sec:supp_cases}}: Qualitative case studies visualizing routing decisions, contrasting efficiency-driven choices in simple OCR tasks against capability-driven choices in complex reasoning scenarios.
    
\end{itemize}

\section{Dataset Details and Composition}
\label{sec:supp_dataset}

Complementing Section 4.2 of the main paper, we provide detailed metadata and preprocessing protocols for all datasets used in MMR-Bench.

\subsection{Dataset Descriptions}
Table \ref{tab:dataset_meta} summarizes the domains, test set sizes, and evaluation metrics for the datasets included in MMR-Bench.

\begin{table}[h]
\centering
\resizebox{\columnwidth}{!}{%
\begin{tabular}{lccc}
\toprule
\textbf{Dataset} & \textbf{Domain} & \textbf{\# Test} & \textbf{Metric} \\
\midrule
OCRBench & OCR & 1,000 & Score \\
SEED-Bench-2-Plus & Text-rich VQA & 2,277 & Accuracy \\
MMStar & VQA & 1,500 & Accuracy \\
RealWorldQA & Real-world QA & 765 & Accuracy \\
MathVista & Visual Math & 1,000 & Accuracy \\
MathVerse & Visual Math & 788 & Accuracy \\
MathVision & Visual Math & 3,040 & Accuracy \\
\bottomrule
\end{tabular}%
}
\caption{Metadata for datasets in MMR-Bench.}
\label{tab:dataset_meta}
\end{table}

\subsection{Data Preprocessing}

We use the \textbf{VLMEvalKit} framework for data processing and evaluation, ensuring standardized inputs tailored to each dataset:

\begin{itemize}
    \item \textbf{Prompts:} Prompts are constructed according to the native format of each dataset:
    \begin{itemize}
        \item For \textbf{Multiple Choice Question (MCQ)} datasets (SEED-Bench-2-Plus, MMStar, RealWorldQA), the input includes the question, candidate options, and an explicit selection instruction (e.g., ``Please select the correct answer from the options above.'').
        \item For \textbf{Visual Question Answering (VQA)} datasets (OCRBench, MathVista, MathVerse, MathVision), we directly use the original question provided by the dataset.
    \end{itemize}
    The final input passed to each MLLM is formatted using its official chat template (e.g., system prompts and role tokens) to match its instruction-tuning distribution.

    \item \textbf{Images:} Images are preprocessed (e.g., resized, interpolated, or padded) following the official implementation and configuration of each candidate MLLM, so that model-specific vision front-ends are respected.
\end{itemize}

Finally, for cost accounting, we log per-instance \textbf{input tokens}, \textbf{image tokens}, and \textbf{output tokens} for each candidate model, and aggregate the corresponding token-level charges under the unified pricing/cost scheme described in Section~\ref{sec:supp_models}.

\section{Model Zoo and Cost Profiling}
\label{sec:supp_models}

\subsection{Candidate Models}
We employ a diverse set of $K=10$ models, encompassing both state-of-the-art proprietary APIs and efficient open-weight models:

\begin{itemize}
    \item \textbf{Open-weight}: InternVL3-78B, Qwen2.5-VL-3B, Qwen2.5-VL-7B, Qwen2.5-VL-72B, Gemma3-4B.
    \item \textbf{Commercial}: GPT-5-0807, GPT-5-Nano-0807, Claude 3.7 Sonnet, Gemini 2.5 Pro, Gemini 2.5 Flash.
\end{itemize}

This model zoo spans a wide range of capacities, architectures, and deployment costs, reflecting the heterogeneity encountered in practical MLLM deployments.

\subsection{Cost Normalization (Eq. 1 \& 3)}
The normalized cost $c_{i}$ is designed to balance commercial API pricing and open-weight inference latency. We normalize the raw cost of each model relative to the most expensive model in the zoo:
\begin{equation}
    c_{i} = \frac{\text{Cost}_{\text{raw}}(i)}{\max(\mathcal{C})},
\end{equation}
where $\mathcal{C}$ is the set of raw costs across all candidate models.

\subsection{Cost Breakdown}
\label{sec:supp_cost_table}

Table \ref{tab:model_cost} lists the \textbf{OpenRouter}\footnote{\url{https://openrouter.ai/} } price for each candidate model, reported as \$/1M \textbf{output} tokens under a unified pricing scheme.

For efficiency-oriented models such as \textbf{GPT-5-Nano-0807} and \textbf{Gemini 2.5 Flash}, we explicitly disable long-form “thinking” or “reasoning” traces during inference to further reduce latency and cost.

\begin{table}[h]
\centering
\caption{Output Token Price per 1M Tokens for Candidate Models (OpenRouter).}
\label{tab:model_cost}
\resizebox{0.85\columnwidth}{!}{%
\begin{tabular}{lcc}
\toprule
\textbf{Model Name} & \textbf{Params / Tier} & \textbf{Price (\$/1M Output Tokens)} \\
\midrule
GPT-5-0807 & High & 10.00 \\
Gemini 2.5 Pro & High & 5.00 \\
Claude 3.7 Sonnet & High & 15.00 \\
Gemini 2.5 Flash & Mid & 2.00 \\
GPT-5-Nano-0807 & Low & 0.50 \\
\midrule
InternVL3-78B & 78B & 0.70 \\
Qwen2.5-VL-72B & 72B & 0.65 \\
Qwen2.5-VL-7B & 7B & 0.07 \\
Gemma3-4B & 4B & 0.04 \\
Qwen2.5-VL-3B & 3B & 0.03 \\
\bottomrule
\end{tabular}%
}
\end{table}

\section{Implementation Details of Routers}
\label{sec:supp_imp}

\subsection{Feature Encoding}
\label{sec:supp_encoding}

All routers operate on frozen instance-level embeddings. Given an input pair $(x_{\text{txt}}, x_{\text{img}})$, we extract:
\begin{itemize}
    \item \textbf{Text embedding.} We encode the question string using the CLIP text encoder with truncation enabled, and $\ell_2$-normalize the resulting vector.
    \item \textbf{Image embedding.} We apply CLIP's official preprocessing transform (resize/crop/normalize), encode it using the CLIP image encoder, and $\ell_2$-normalize the resulting vector.
\end{itemize}

\subsection{Feature Fusion Mechanism}
For multimodal routing, we employ a parameter-free, adaptive weighting mechanism to fuse textual and visual embeddings. Let $x_{\text{txt}}, x_{\text{img}} \in \mathbb{R}^d$ denote the $\ell_2$-normalized text and image embeddings for instance $i$. The fused representation $z_i$ is computed in three steps:

\begin{figure*}[t]
    \centering
    \begin{subfigure}[b]{0.32\textwidth}
        \centering
        \includegraphics[width=\textwidth]{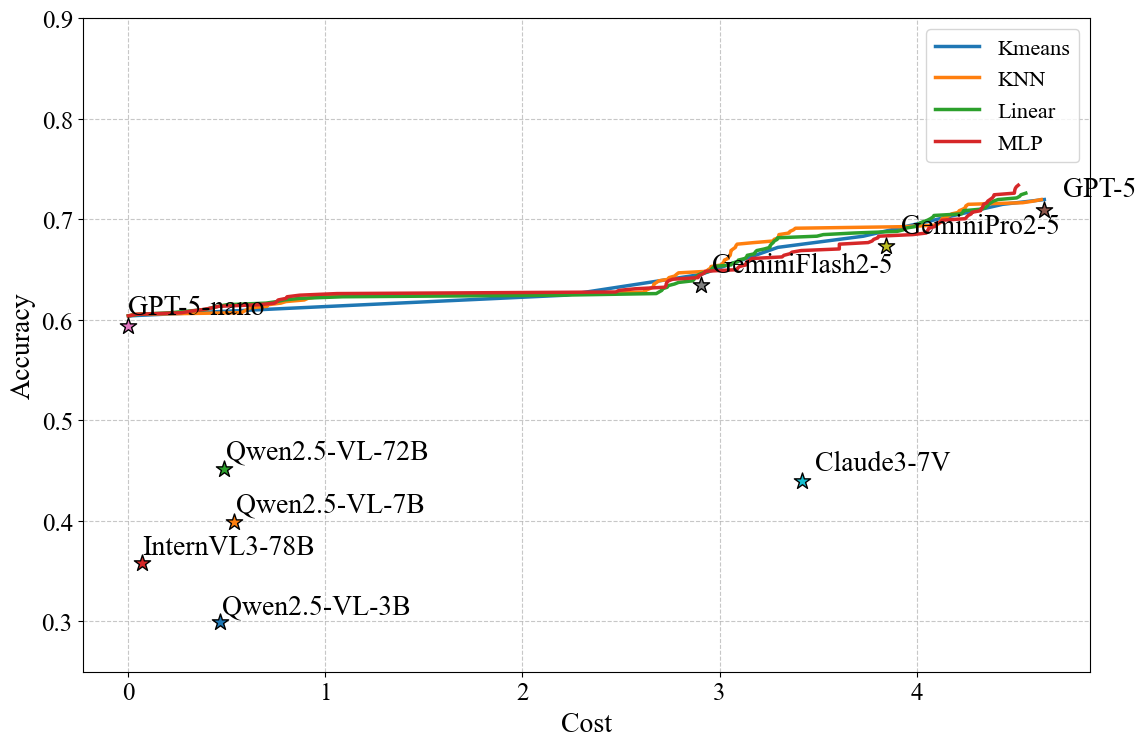}
        \caption{MathVista}
        \label{fig:res_mathvista}
    \end{subfigure}
    \hfill
    \begin{subfigure}[b]{0.32\textwidth}
        \centering
        \includegraphics[width=\textwidth]{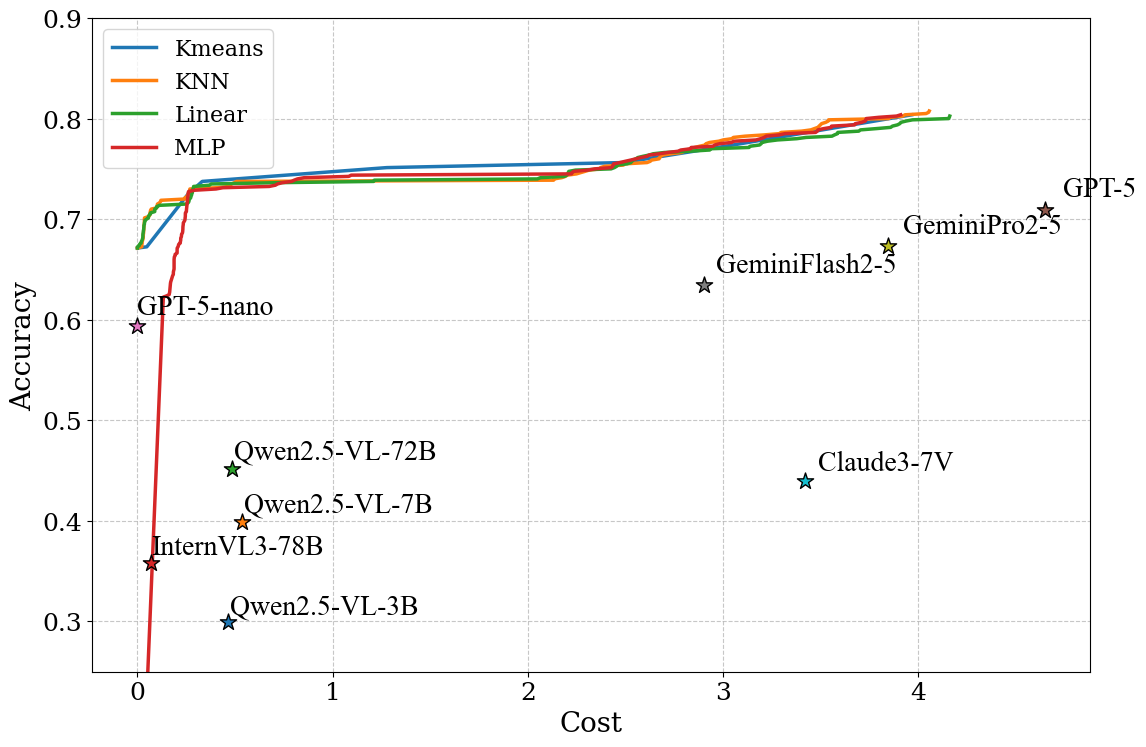}
        \caption{MathVerse}
        \label{fig:res_mathverse}
    \end{subfigure}
    \hfill
    \begin{subfigure}[b]{0.32\textwidth}
        \centering
        \includegraphics[width=\textwidth]{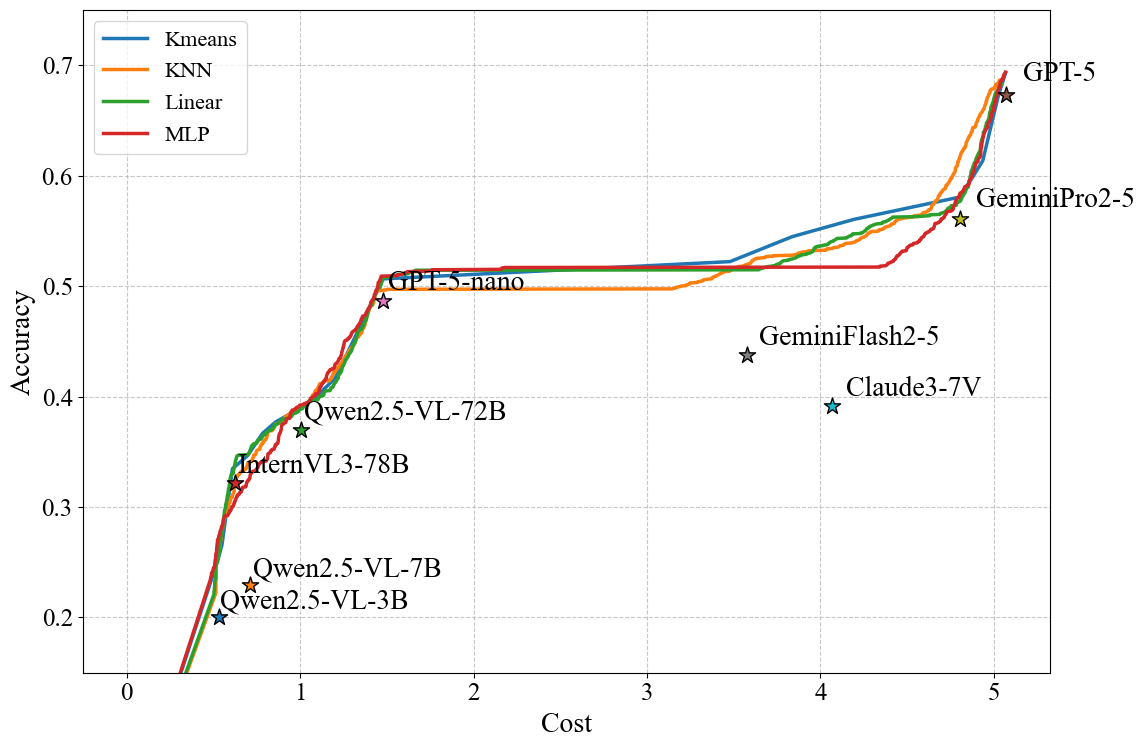} 
        \caption{MathVision}
        \label{fig:res_mathvision}
    \end{subfigure}
    
    \vspace{0.3cm} 

    \begin{subfigure}[b]{0.32\textwidth}
        \centering
        \includegraphics[width=\textwidth]{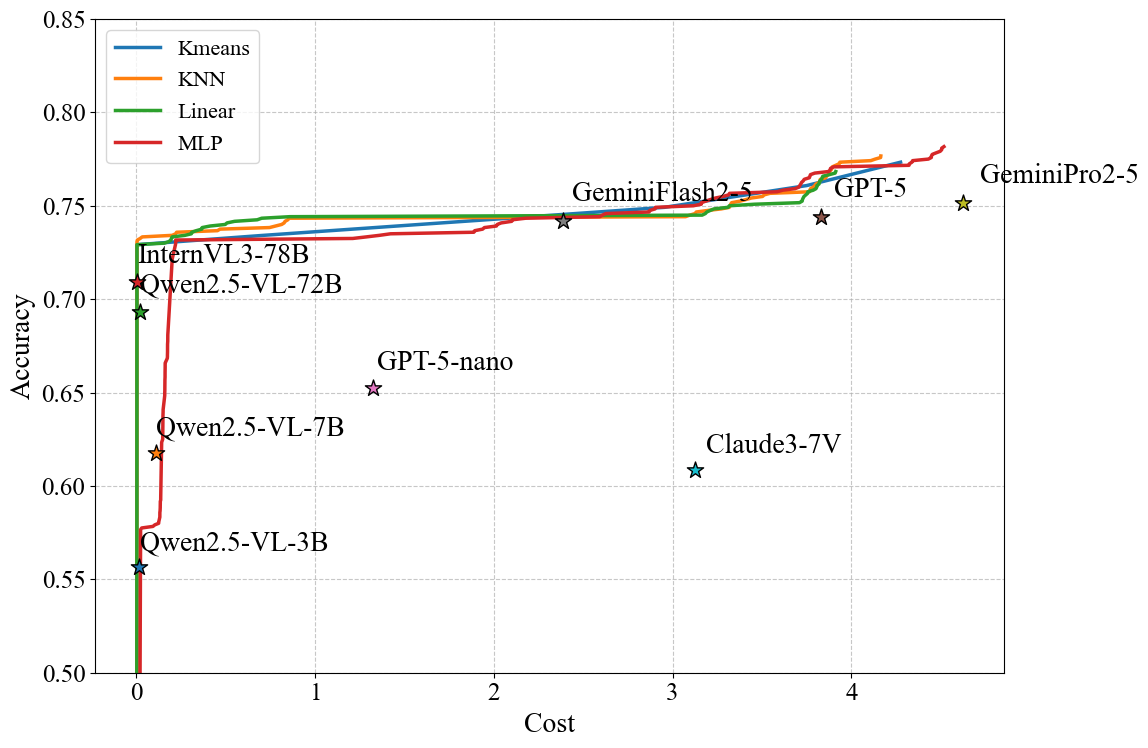}
        \caption{MMStar}
        \label{fig:res_mmstar}
    \end{subfigure}
    \hfill
    \begin{subfigure}[b]{0.32\textwidth}
        \centering
        \includegraphics[width=\textwidth]{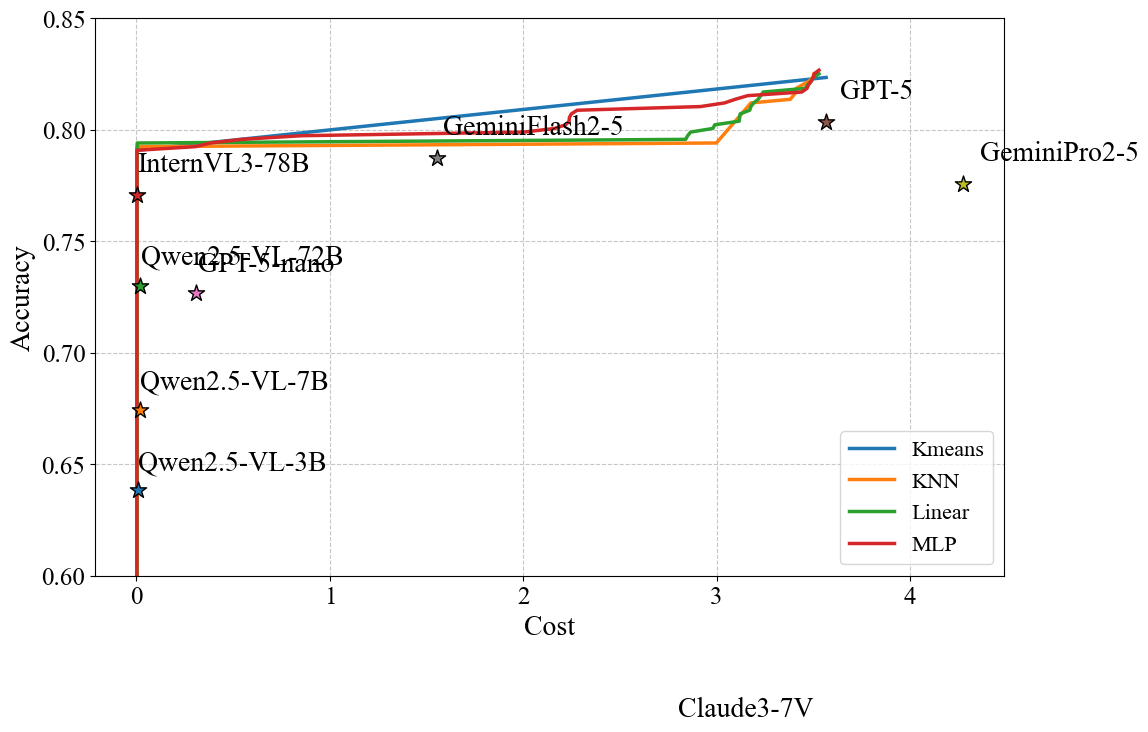}
        \caption{RealWorldQA}
        \label{fig:res_rwqa}
    \end{subfigure}
    \hfill
    \begin{subfigure}[b]{0.32\textwidth}
        \centering
        \includegraphics[width=\textwidth]{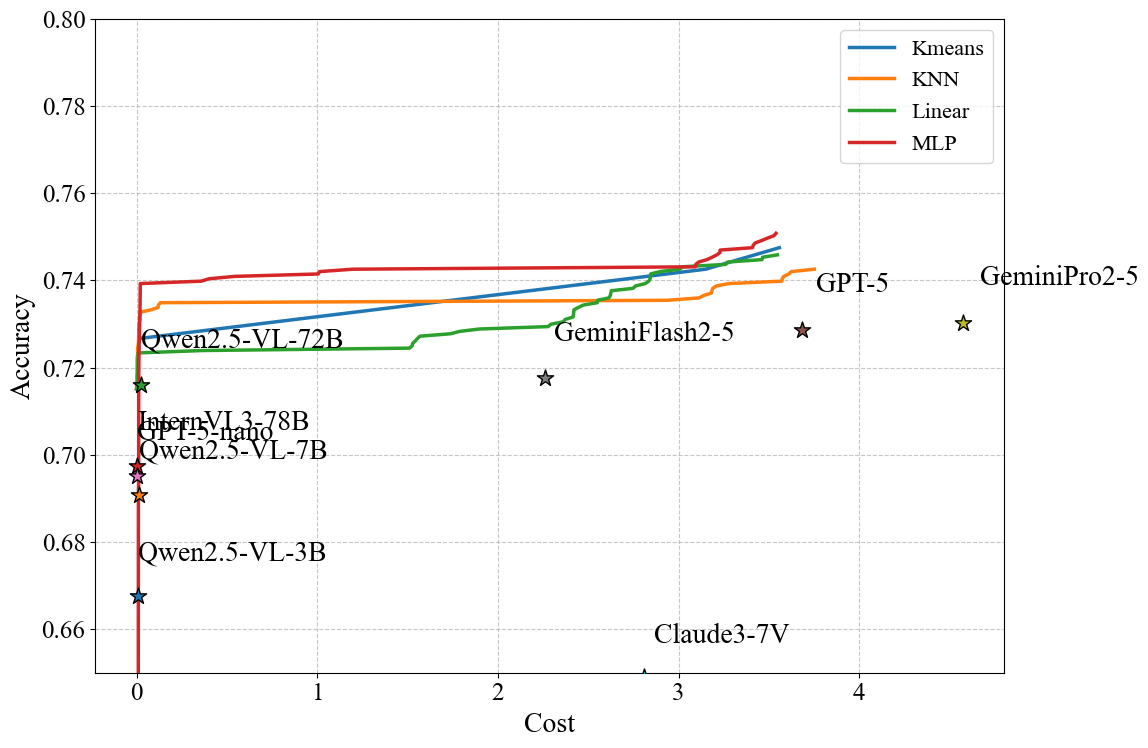}
        \caption{SEED-Bench}
        \label{fig:res_seedbench}
    \end{subfigure}

    \caption{\textbf{Cost-Accuracy Pareto Frontiers across Six Datasets.} 
    The curves illustrate the trade-off between normalized cost (x-axis) and accuracy (y-axis). 
    Routing strategies (colored lines) consistently envelop the single-model baselines (stars), demonstrating superior efficiency.
    The top row represents mathematical reasoning tasks, while the bottom row covers general VQA and OCR scenarios.}
    \label{fig:six_dataset_breakdown}
\end{figure*}

\begin{enumerate}
    \item \textbf{Confidence Estimation.} We compute a lightweight confidence score for each modality by combining:
    (i) \emph{prototype similarity} (cosine similarity to the modality mean embedding), and
    (ii) a \emph{norm-based} signal (how large the embedding norm is relative to the current encoded set).
    Concretely, define modality prototypes $\mu_{\text{txt}}=\frac{1}{n}\sum_i x_{\text{txt}}$ and $\mu_{\text{img}}=\frac{1}{n}\sum_i x_{\text{img}}$. For each instance $i$:
    \begin{equation}
    \begin{aligned}
        s_{\text{txt}}^{(i)} &= \cos(x_{\text{txt}}^{(i)}, \mu_{\text{txt}}), \quad
        s_{\text{img}}^{(i)} = \cos(x_{\text{img}}^{(i)}, \mu_{\text{img}}),\\
        c_{\text{txt,proto}}^{(i)} &= \frac{s_{\text{txt}}^{(i)} + 1}{2}, \quad
        c_{\text{img,proto}}^{(i)} = \frac{s_{\text{img}}^{(i)} + 1}{2}.
    \end{aligned}
    \end{equation}
    Let $r_{m}^{(i)}=\|x_{m}^{(i)}\|_2$. With mean and standard deviation over the current encoded set $(\bar r_m,\sigma_m)$, we map the standardized norm to $(0,1)$ via a sigmoid:
    \begin{equation}
        c_{m,\text{norm}}^{(i)}=\sigma\!\left(\frac{r_m^{(i)}-\bar r_m}{\sigma_m}\right), \quad \sigma(t)=\frac{1}{1+e^{-t}}.
    \end{equation}
    Finally, we combine the two signals with equal weights and clip to $[0,1]$ for numerical stability:
    \begin{equation}
        c_m^{(i)} = \mathrm{clip}\Big(0.5\cdot c_{m,\text{proto}}^{(i)} + 0.5\cdot c_{m,\text{norm}}^{(i)},\, 0,\, 1\Big).
    \end{equation}

    \item \textbf{Adaptive Weighting.} We convert these confidence scores into soft weights via a softmax with temperature $\eta = 5.0$:
    \begin{equation}
        w_{\text{txt}}^{(i)}, w_{\text{img}}^{(i)} = \text{softmax}\big(\eta \cdot [c_{\text{txt}}^{(i)}, c_{\text{img}}^{(i)}]\big).
    \end{equation}
    \item \textbf{Interaction Terms:} To capture cross-modal dynamics, we explicitly include element-wise product and difference terms:
    \begin{equation}
        z_i = \Big(w_{\text{txt}}^{(i)} x_{\text{txt}}^{(i)} + w_{\text{img}}^{(i)} x_{\text{img}}^{(i)}\Big) + \alpha \cdot \big(x_{\text{txt}}^{(i)} \odot x_{\text{img}}^{(i)}\big) + \beta \cdot \big|x_{\text{txt}}^{(i)} - x_{\text{img}}^{(i)}\big|.
    \end{equation}
    The final vector $z_i$ is $\ell_2$-normalized before being fed into the router.
\end{enumerate}

\subsection{Router Architectures and Hyperparameters}
\label{sec:supp_router_arch}

\begin{table}[t]
\centering
\caption{\textbf{Router summary in our implementation.} ``Fusion feature'' specifies how text/image embeddings are combined before routing.}
\label{tab:router_hparams}
\resizebox{\columnwidth}{!}{%
\begin{tabular}{lll}
\toprule
\textbf{Router} & \textbf{Type} & \textbf{Fusion feature} \\
\midrule
KMeansRouter & Non-parametric & Adaptive fusion \\
KNNRouter & Non-parametric & Adaptive fusion \\
LinearRouter & Regression & Average fusion \\
MLPRouter & Regression & Average fusion \\
LinearMFRouter & Low-rank regression & Adaptive fusion + SVD \\
MLPMFRouter & MF-style NN & Adaptive fusion + latent factors \\
CrossModalRouter (CMR) & Cross-modal attention & Average fusion (compat) \\
RandomRouter & Baseline & None \\
OracleRouter & Upper bound & None \\
\bottomrule
\end{tabular}%
}
\end{table}

\paragraph{Prediction targets and selection rule.}
For a test instance $i$ and each candidate model $j$, routers predict a utility $\hat{u}_{i,j}$ (higher is better) and a cost $\hat{c}_{i,j}$ (lower is better).
Given a trade-off weight $\lambda\ge 0$, we select
\begin{equation}
    j^\star(i;\lambda)=\arg\min_j \big(1-\hat{u}_{i,j}+\lambda\,\hat{c}_{i,j}\big),
\end{equation}
and then report the \emph{true} utility/cost of the chosen model from the offline outcome table.
Unless otherwise stated, utility and cost predictors are trained independently, and NaNs are handled conservatively (e.g., masked in the loss, or excluded from averaging in neighbor/cluster aggregation).

\subsubsection{Non-Parametric Routers}

\noindent \textbf{KMeansRouter.}
We cluster training instances in the fused embedding space using K-Means++ and represent each cluster by the per-model average utility and cost of its members.
With $C$ clusters and assignments $h(i)\in\{1,\ldots,C\}$, we compute NaN-safe empirical means:
\begin{equation}
    \hat{u}_{h,j}=\mathbb{E}[u_{i,j}\mid h(i)=h],\quad
    \hat{c}_{h,j}=\mathbb{E}[c_{i,j}\mid h(i)=h].
\end{equation}
At inference time, we assign each test instance to its nearest cluster centroid (in the fused embedding space) and set
$(\hat{u}_{i,j},\hat{c}_{i,j})=(\hat{u}_{h(i),j},\hat{c}_{h(i),j})$.

\vspace{2mm}
\noindent \textbf{KNNRouter.}
We retrieve the $k$ nearest neighbors in the fused embedding space and estimate per-model utility/cost by neighbor averaging.
We use \textbf{NearestNeighbors} with cosine distance.
For each test instance $i$ and model $j$:
\begin{equation}
\begin{aligned}
\hat{u}_{i,j} &= \mathrm{mean}\{u_{\ell,j}:\ell\in\mathcal{N}_k(i)\},\\
\hat{c}_{i,j} &= \mathrm{mean}\{c_{\ell,j}:\ell\in\mathcal{N}_k(i)\}.
\end{aligned}
\end{equation}
\subsubsection{Regression Routers}

\noindent \textbf{LinearRouter.}
We fit two independent multi-output linear regressors to map embeddings to per-model utility and cost.
For multimodal inputs, we use a simple equal-weight average of text and image embeddings:
\begin{equation}
    x_{\text{avg}}=\tfrac{1}{2}\big(x_{\text{txt}}+x_{\text{img}}\big).
\end{equation}

\noindent \textbf{MLPRouter.}
We replace linear regression with two independent 2-layer MLPs (one for utility and one for cost).
The architecture is Linear--ReLU--Linear--ReLU--Linear.
We train with MSE loss using Adam.
The multimodal input feature is the same averaged embedding $x_{\text{avg}}$ as above.

\subsubsection{Low-Rank and MF-Style Routers}

\noindent \textbf{LinearMFRouter (low-rank regression).}
This router combines adaptive fusion (Section~\ref{sec:supp_imp}) with a shared low-rank projection and ridge regression.
Given fused features $z_i\in\mathbb{R}^d$, we first project them to rank $r$ with \textbf{TruncatedSVD}, then train two multi-output ridge regressors:
\begin{equation}
    \tilde{z}_i=\mathrm{SVD}(z_i)\in\mathbb{R}^r,\quad
    \hat{u}_{i,\cdot}=\mathrm{Ridge}(\tilde{z}_i),\quad
    \hat{c}_{i,\cdot}=\mathrm{Ridge}(\tilde{z}_i).
\end{equation}

\noindent \textbf{MLPMFRouter (matrix-factorization style).}
We learn a shared latent space for instances and candidate models.
A feature network maps each instance embedding to a latent factor $z_{\text{lat}}^{(i)}\in\mathbb{R}^{r}$, while each model $j$ has a learned vector $w_j\in\mathbb{R}^r$:
\begin{equation}
    z_{\text{lat}}^{(i)}=\mathrm{MLP}(z_i),\quad
    \hat{u}_{i,j}=(z_{\text{lat}}^{(i)})^\top w_j + b_j,
\end{equation}
and analogously for costs using a separate feature network and model vectors.
Training minimizes an observed-entry MSE:
\begin{equation}
    \mathcal{L}=\frac{1}{|\Omega|}\sum_{(i,j)\in\Omega}\big(\hat{u}_{i,j}-u_{i,j}\big)^2,
\end{equation}
with an independent loss for costs.

\subsubsection{Cross-Modal Attention Router}

\textbf{CrossModalRouter (CMR).}
We also include an attention-based baseline that models cross-modal interactions using stacked multi-head attention layers over modality tokens.
It trains \emph{per-expert} predictors: a binary classifier for correctness (BCE with logits, with optional class balancing via \textbf{pos\_weight}) and a regressor for cost (MSE), optionally predicting $\log(1+c)$ to reduce skew.

\subsubsection{Baselines}

\noindent \textbf{RandomRouter.}
Selects a model uniformly at random for each instance; no training is performed.

\noindent \textbf{OracleRouter.}
An upper bound that selects using ground-truth utilities/costs on the test split.
In the cost-unaware form, it chooses $\arg\max_j u_{i,j}$; in the cost-aware form, it minimizes $(1-u_{i,j}+\lambda c_{i,j})$.

\section{Additional Experimental Results}
\label{sec:supp_exp}

\subsection{Quality-Neutral Cost (QNC) Analysis}

To complement accuracy, we further compare the cost–effectiveness of different routers using
the Quality-Neutral Cost (QNC) metric. Table~\ref{tab:qnc_router} reports QNC for each router
across all eight datasets, as well as the average over all benchmarks (\emph{All}). Recall that
QNC $<1$ indicates that the router reaches the target accuracy at a lower cost than the best
single model, while QNC $=\infty$ means the router cannot reach the target accuracy even when
routing over all candidate models.

\begin{table*}[t]
\centering
\caption{\textbf{Quality-Neutral Cost (QNC) across routers and datasets.}
QNC $<1$ indicates that the router achieves the target accuracy at a lower cost than the best single model.
$\infty$ denotes that the router cannot match the target accuracy even when routing over all candidate models.}
\label{tab:qnc_router}
\resizebox{\textwidth}{!}{%
\begin{tabular}{lcccccccc}
\toprule
\multirow{2}{*}{\textbf{Router}} & \multicolumn{8}{c}{$\textbf{QNC} \downarrow$} \\
\cmidrule(lr){2-9}
 & OCRBench & SEED-Bench-2-Plus & MathVerse & MathVista & MathVision & MMStar & RealWorldQA & All \\
\midrule
KMeans   & 1.000 & $\infty$ & 1.000 & 1.000 & 1.000 & 0.702 & 1.000 & 1.059 \\
MLPMF    & 1.000 & 0.344    & 0.757 & 0.944 & 0.998 & 0.737 & 0.934 & 0.995 \\
LinearMF & 0.984 & $\infty$ & 0.789 & 0.948 & $\infty$ & $\infty$ & 0.934 & 0.970 \\
KNN      & 0.987 & $\infty$ & 0.984 & 0.965 & $\infty$ & 0.486 & 0.931 & $\infty$ \\
Oracle   & 0.920 & $2.55\times 10^{-5}$ & $2.61\times 10^{-4}$ & $4.20\times 10^{-4}$ & $1.58\times 10^{-2}$ & $3.61\times 10^{-5}$ & $1.09\times 10^{-4}$ & $2.62\times 10^{-3}$ \\
\bottomrule
\end{tabular}%
}
\end{table*}

Overall, the learnable MF-based routers (MLPMF and LinearMF) achieve QNC $<1$ on several
benchmarks (e.g., SEED-Bench-2-Plus, MathVerse, MMStar), and also obtain competitive average
QNC on \emph{All} ($0.995$ and $0.970$), indicating that they are at least as cost-efficient
as deploying the best single model. In contrast, the non-parametric KMeans baseline remains
above $1$ on average, and KNN suffers from QNC $=\infty$ on multiple datasets, highlighting
the benefit of learnable fusion and routing. As expected, the Oracle provides a strong but
impractical upper bound with near-zero QNC.

\subsection{Detailed Analysis of Pareto Frontiers}

Figure \ref{fig:six_dataset_breakdown} visualizes the performance-cost trade-offs across six diverse datasets. 
Across all scenarios, the learned routing policies consistently form a convex hull above the single-model baselines, validating that dynamic model selection yields a strictly better Pareto frontier than any static choice.
In mathematical reasoning tasks (Top Row), we observe that instance-based methods (e.g., KNN, orange line) often exhibit sharper performance jumps at lower costs, suggesting that mathematical problems possess high structural repeatability that retrieval-based routing effectively exploits. 
Conversely, in general VQA and OCR benchmarks (Bottom Row), parametric routers (e.g., Linear/MLP, green/red lines) demonstrate smoother and more robust scaling, effectively bridging the significant cost gap between efficient open-weight models (e.g., Qwen2.5-VL) and high-capability proprietary APIs (e.g., GPT-5) by learning global difficulty distributions. 
Notably, on datasets like SEED-Bench and MMStar, the routers achieve near-optimal accuracy while activating expensive models for only a small fraction of hard queries, thereby minimizing the average inference cost.

\begin{figure*}[t]
    \centering

    \begin{subfigure}[b]{\textwidth}
        \centering
        \begin{tcolorbox}[colback=white, colframe=gray!20, boxrule=1pt, arc=2mm, width=\textwidth, halign=center]
            \includegraphics[height=3.5cm, keepaspectratio]{fig/sup/20.jpg}
        \end{tcolorbox}
        \vspace{-0.2cm}
        \begin{tcolorbox}[colback=green!5, colframe=green!40, boxrule=1pt, arc=2mm, width=\textwidth, title=\textbf{\small Case A: Efficiency in OCR}]
            \small
            \textbf{Query:} ``What is written in the image?'' \\
            \textbf{Router Decision:} \textcolor{green!50!black}{\textbf{Qwen2.5-VL-3B}} (Cost $\approx$ 0) \\
            \textbf{Outcome:} Correct (``motorsports'') \\
            \textbf{Savings:} \textbf{95\% cost reduction} vs. GPT-5.
        \end{tcolorbox}
    \end{subfigure}

    \vspace{0.35cm}

    \begin{subfigure}[b]{\textwidth}
        \centering
        \begin{tcolorbox}[colback=white, colframe=gray!20, boxrule=1pt, arc=2mm, width=\textwidth, halign=center]
            \includegraphics[height=3.5cm, keepaspectratio]{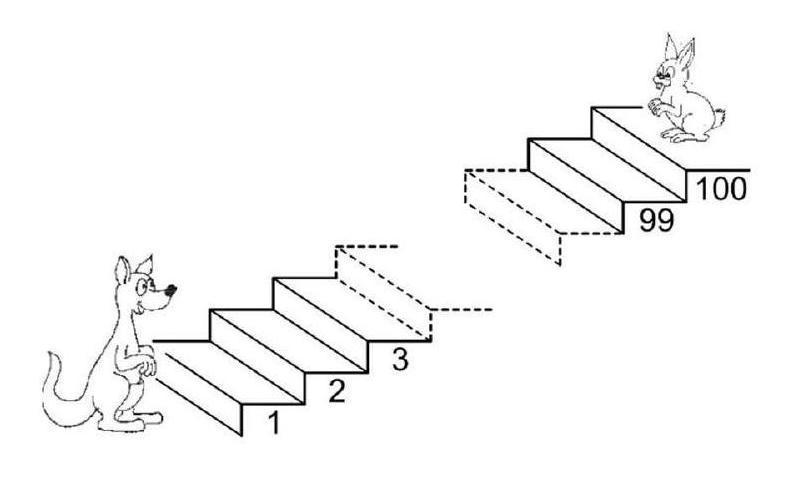}
        \end{tcolorbox}
        \vspace{-0.2cm}
        \begin{tcolorbox}[colback=blue!5, colframe=blue!40, boxrule=1pt, arc=2mm, width=\textwidth, title=\textbf{\small Case B: Multimodal Reasoning}]
            \small
            \textbf{Query:} ``...Whenever the kangaroo goes up seven steps, the rabbit goes down three...'' \\
            \textbf{Router Decision:} \textcolor{red!60!black}{\textbf{GPT-5}} (Max Capability) \\
            \textbf{Outcome:} Correct (``76'') \\
            \textbf{Insight:} Smaller models failed to ground the initial position or hallucinated the calculation.
        \end{tcolorbox}
    \end{subfigure}

    \caption{\textbf{Qualitative Visualization of Routing Decisions.}
    \textbf{(Top)} For a clear OCR task, the router identifies low difficulty and dispatches the query to the most cost-effective model (Qwen2.5-VL-3B), avoiding unnecessary expense.
    \textbf{(Bottom)} For a complex mathematical word problem requiring visual grounding and multi-step logic, the router correctly detects the high capability requirement and invokes the strongest model (GPT-5), ensuring accuracy.}
    \label{fig:qualitative_cases}
\end{figure*}

\section{Qualitative Case Studies}
\label{sec:supp_cases}

To better understand the decision boundaries of our learned routers, we visualize two representative cases in Figure \ref{fig:qualitative_cases}.

\textbf{Case A (Efficiency via Routing):} The input image contains large, unambiguous text ("MOTORSPORTS"). The router's fusion module detects strong alignment between the visual text features and the simple extraction query. Consequently, it assigns the task to \textbf{Qwen2.5-VL-3B}, the smallest model in the zoo. The model answers correctly, demonstrating that for routine perception tasks, "heavy" models like GPT-5 are often over-provisioned. The router effectively prunes this redundancy.

\textbf{Case B (Capability via Routing):} The second case involves a "Kangaroo and Rabbit" math problem. This query is deceptively simple in text but requires precise visual grounding (identifying the Rabbit starts at step 100) and multi-step logical reasoning (calculating the number of jumps and the corresponding descent). The router recognizes the complexity cues—likely triggered by the length of the reasoning chain required and the diagrammatic nature of the image—and correctly routes the query to \textbf{GPT-5}. While smaller models struggled with either the initial state extraction or the arithmetic logic, the routed system achieved the correct answer ("76").

\textbf{Conclusion:} These examples highlight the core value proposition of MMR-Bench: distinguishing between "commodity" tasks that can be solved cheaply and "premium" tasks that demand frontier capabilities, thereby optimizing the return on computational investment.


\end{document}